\documentclass[runningheads]{llncs}
\usepackage{diagbox}
\usepackage{xcolor}
\usepackage{siunitx}
\usepackage{svg}
\usepackage{booktabs}
\usepackage{caption}
\usepackage{adjustbox}
\usepackage{amsmath}
\usepackage{multirow}
\usepackage{array}
\usepackage[T1]{fontenc}
\usepackage{siunitx}
\usepackage{fontawesome5}
\usepackage{graphicx}
\usepackage{amsmath}
\usepackage{amsfonts}
\graphicspath{ {images/} }  
\usepackage{caption}  
\usepackage[colorlinks=true, allcolors=blue]{hyperref} 
\usepackage[capitalise]{cleveref} 
\usepackage{multirow}

\begin{document}
\title{Enhancing Fundus Image-based Glaucoma Screening via Dynamic Global-Local Feature Integration}
%
%

\author{Yuzhuo Zhou\inst{1} \and
Chi Liu \inst{2}\faIcon{envelope} \and
Sheng Shen \inst{3} \and
Siyu Le \inst{1} \and
Liwen Yu \inst{1} \and
Sihan Ouyang \inst{1} \and
Zongyuan Ge \inst{4}
}
\authorrunning{Y. Zhou et al.}
%
\institute{Minzu University Of China, China \and
Faculty of Data Science, City University of Macau, Macao SAR, China \and
Design and Creative Technology Vertical, Torrens University Australia, NSW, Australia \and 
Faculty of Information Technology, Monash University, Melbourne, VIC, Australia \\
\faIcon{envelope} Corresponding author: \email{chiliu@cityu.edu.mo}
}

\maketitle              
\begin{abstract}
With the advancements in medical artificial intelligence (AI), fundus image classifiers are increasingly being applied to assist in ophthalmic diagnosis. 
While existing classification models have achieved high accuracy on specific fundus datasets, they struggle to address real-world challenges such as variations in image quality across different imaging devices, discrepancies between training and testing images across different racial groups, and the uncertain boundaries due to the characteristics of glaucomatous cases. 
In this study, we aim to address the above challenges posed by image variations by highlighting the importance of incorporating comprehensive fundus image information, including the optic cup (OC) and optic disc (OD) regions, and other key image patches. 
Specifically, we propose a self-adaptive attention window that autonomously determines optimal boundaries for enhanced feature extraction.
Additionally, we introduce a multi-head attention mechanism to effectively fuse global and local features via feature linear readout, improving the model’s discriminative capability. 
Experimental results demonstrate that our method achieves superior accuracy and robustness in glaucoma classification. 

\end{abstract}

\section{Introduction}
Glaucoma is a major ocular pathology and one of the leading causes of irreversible blindness worldwide. 
Early screening is an effective strategy for detecting glaucomatous alteration in its initial stages, facilitating timely intervention before substantial visual impairment occurs. 
Fundus photography plays a key role in early glaucoma screening due to its non-invasive nature and cost-effectiveness. 
The recent integration of AI into automated fundus image analysis has demonstrated remarkable potential in this field, achieving accuracy levels comparable to or even surpassing those of human ophthalmologists. 
This advancement holds substantial promise for improving global eye care systems and is increasingly being adopted in clinical practice.

\begin{figure}
    \centering
    \includegraphics[width=\linewidth]{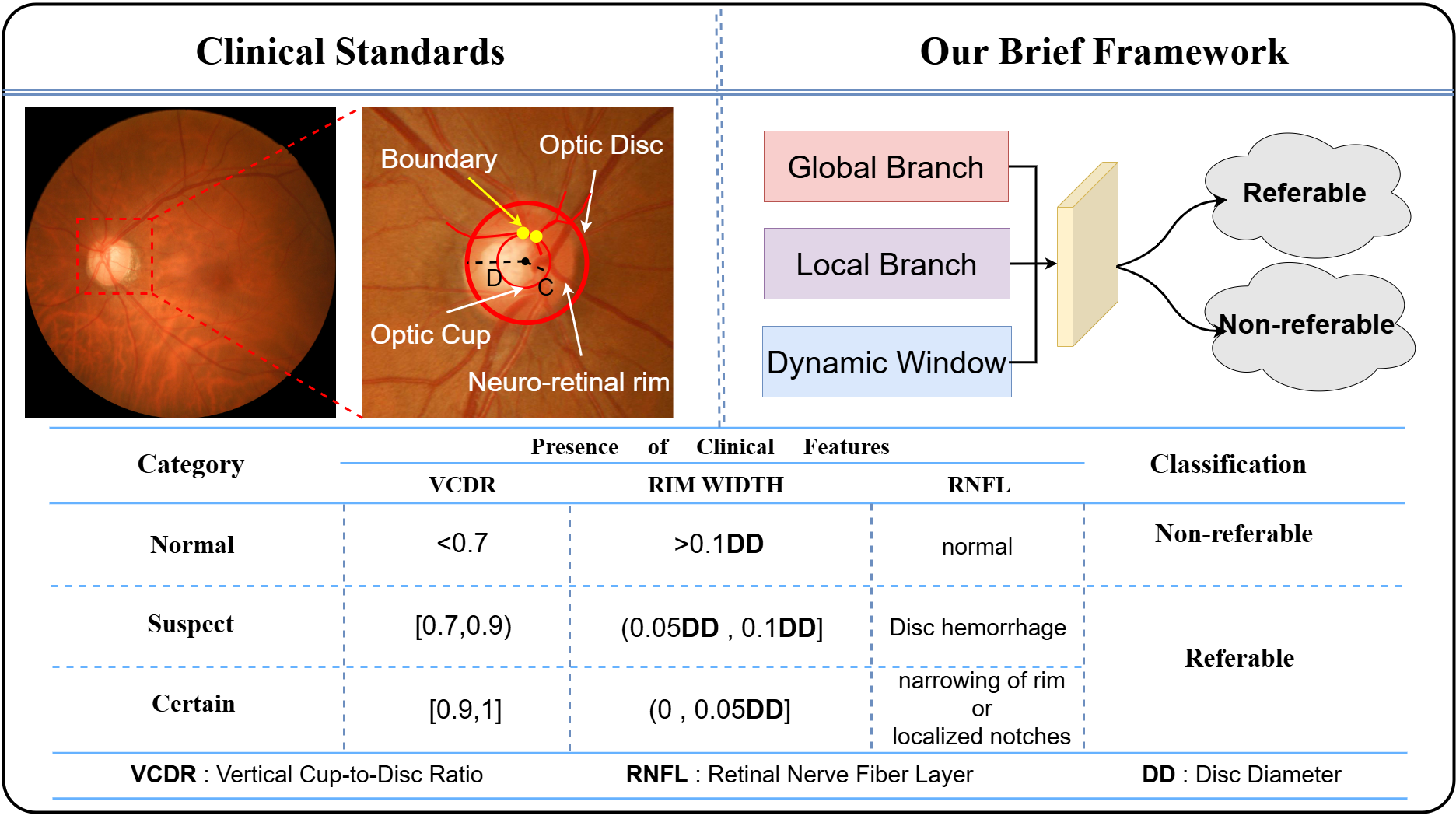}
    \caption{The motivation of our method: Detecting glaucoma with aligning with clinical standards.}
    \label{fig:enter-label}
\end{figure}

In glaucoma screening, traditional methods primarily rely on the visual assessment of fundus images by clinicians. This process is highly subjective, as it is heavily influenced by the clinicians' individual background knowledge and clinical experience. As shown in the \cref{fig:enter-label}, the delineation of the cup-disc boundary is of critical importance, and is commonly identified at the first bend of small blood vessels within the OC. 
However, this cup-disc boundary does not have fixed, predefined criteria, as the locations vary across the diverse structures of the small blood vessels among patients.
This delineation is entirely subjective. 
Such subjectivity extends beyond cup-disc boundary delineation to multiple aspects of the diagnostic process, including the identification of subtle ocular lesions in fundus images, variations in imaging quality, and the assessment of lesion severity. 
These subjective factors contribute to substantial inconsistencies in segmenting criteria and glaucoma diagnosis, undermining the reliability of the overall screening process.

To mitigate clinician subjectivity and enhance the accuracy of glaucoma diagnosis, there have been increasing attempts to apply AI for fundus image-based glaucoma screening.
These AI methods classify patients as 'referable' or 'non-referable' based on the detected disease status in their fundus images.
The primary objective is to enable AI to capture robust pathological features associated with early-stage glaucoma through a deep feature encoder. 
As illustrated in \cref{fig:enter-label}, the clinical criteria incorporates early indicators of the disease, including small splinter hemorrhages on the OD and defects in the retinal nerve fiber layer, which are typically located at the superotemporal or inferotemporal margin.
However, these margins do not have fixed predefined boundaries, as their locations vary across cases. 
A major limitation of existing deep encoders is their inability to account for the natural variability present in real-world data\cite{r13}, thereby restricting their accuracy and efficiency in handling diverse image qualities and fused datasets\cite{r10}. These deep encoders lack a dynamic mechanism to adaptively determine the optimal receptive region for accurate defect detection.
Moreover, most approaches\cite{r8,r9} rely on global fundus images as input, often overlooking local details critical to detecting glaucomatous alterations, which primarily manifest around the OC and OD. 
Including irrelevant regions can increase the encoder's susceptibility to global imaging noise, such as overexposure and shadows. 
Nevertheless, global imaging noise is less prevalent in the OC and OD areas, whereas the presence of co-existing ocular diseases with pathological defects outside these regions may further degrade the encoder’s performance.
Therefore, effectively integrating global and local branches to develop a robust deep feature representation that generalizes across various image qualities and datasets warrants further exploration.

To address the inherent limitations of deep encoders, we propose a cross-attention three-branch model. Our model can capture the inherent fuzziness of the cup-disc boundary, reflecting the uncertainty in clinical decision-making.
The model consists of three branches: the global branch, the local branch, and the dynamic window mechanism (DWM)-based local branch.
The global branch captures spatial features at a global level, while the local branch extracts features from the region of interest (ROI).
The DWM, as the core component of our model, automatically selects the optimal receptive fields across entire images. Similar to the second branch, it belongs to the local branches but introduces a dynamic selection process. The optimal receptive fields are determined by computing the maximum and total scores of each feature map. Using these scores, we compute the centers of selected patches and determine the boundaries of the selected fields based on the scales of the receptive regions, which are localized through the top-left and bottom-right corner indices of the selected receptive patch.
Moreover, we integrate a convolutional block attention mechanism (CBAM) \cite{woo2018cbam} to mitigate the information redundancy inherent in traditional convolutional neural networks. 
By leveraging the channel attention and the spatial attention, our model not only enhances feature extraction across various channels and spatial regions but also minimizes the impact of irrelevant features in the surrounding regions identified by DWM. 
By incorporating DWM and CBAM, the third branch effectively mitigates the uncertainty in clinical margin decision-making.
The experiments demonstrate that our model reduces reliance on individual clinical experience, fostering a more objective and robust determination of the cup-disc boundary and significantly improving the accuracy of glaucoma diagnosis.
Our main contributions are summarized as follows:

\begin{enumerate}
    \item We propose a cross-attention three-branch model to capture the inherent fuzziness of the cup-disc boundary in glaucoma screening, addressing the uncertainty in clinical decision-making. 
    
    \item We introduce DWM into a patch selection module that autonomously identifies the optimal receptive fields using a dynamic window, minimizing the likelihood of focusing on irrelevant information and improving feature localization.

    \item We integrate CBAM to enhance the feature extractor's ability to select deep relevant features. This mechanism ensures our approach's performance and robustness, guaranteeing consistent effectiveness across diverse model architectures.

\end{enumerate}

\section{RELATED WORK}
Accurate Glaucoma detection is becoming increasingly crucial in automated ophthalmologic diagnosis. 
Recently, deep learning techniques have been widely adopted\cite{r1} to improve model's performance in global and local feature fusion. To enhance the ability to extract local features, models have been optimized for segmenting the OC and OD, capturing deeper local features and mitigating deviations caused by the diversity of unseen datasets.
Li et al.\cite{r25} improved model performance in OC and OD segmentation by utilizing a disc proposal network and a cup proposal network in conjunction with an end-to-end region-based deep convolutional neural network.
Similarly, Huang et al.\cite{r9} proposed a dynamic-local learning module incorporating deformable convolution, which enhances the ability to focus on local features from low-resolution medical images. Xu et al.\cite{xu2012efficient} further designed interest mechanisms to localize the OC and OD.

Due to the complexity of fundus feature extraction, attention-based mechanism have gained increasing attention within the academic community.
Salam et al.\cite{r26} designed an autonomous glaucoma detection algorithm that integrates structural and non-structural features using machine learning. Sinthanayothin et al.\cite{r30} utilized color contrast to enhance the capacity for OD localization. 
Guo et al.\cite{r27} proposed a neural network named CP-FD-UNet++, which incorporates input and feature maps at different scales, while their proposed IFOV model extracts hidden visual features from the gray-level co-occurence matrix. However, since Guo et al.'s method addresses multiple scales of feature maps, it does not specifically focus on receptive field localization.

\section{METHOD}
\subsection{Overall Of The Method}
\begin{figure}[htbp]
    \centering
    \includegraphics[width=\linewidth]{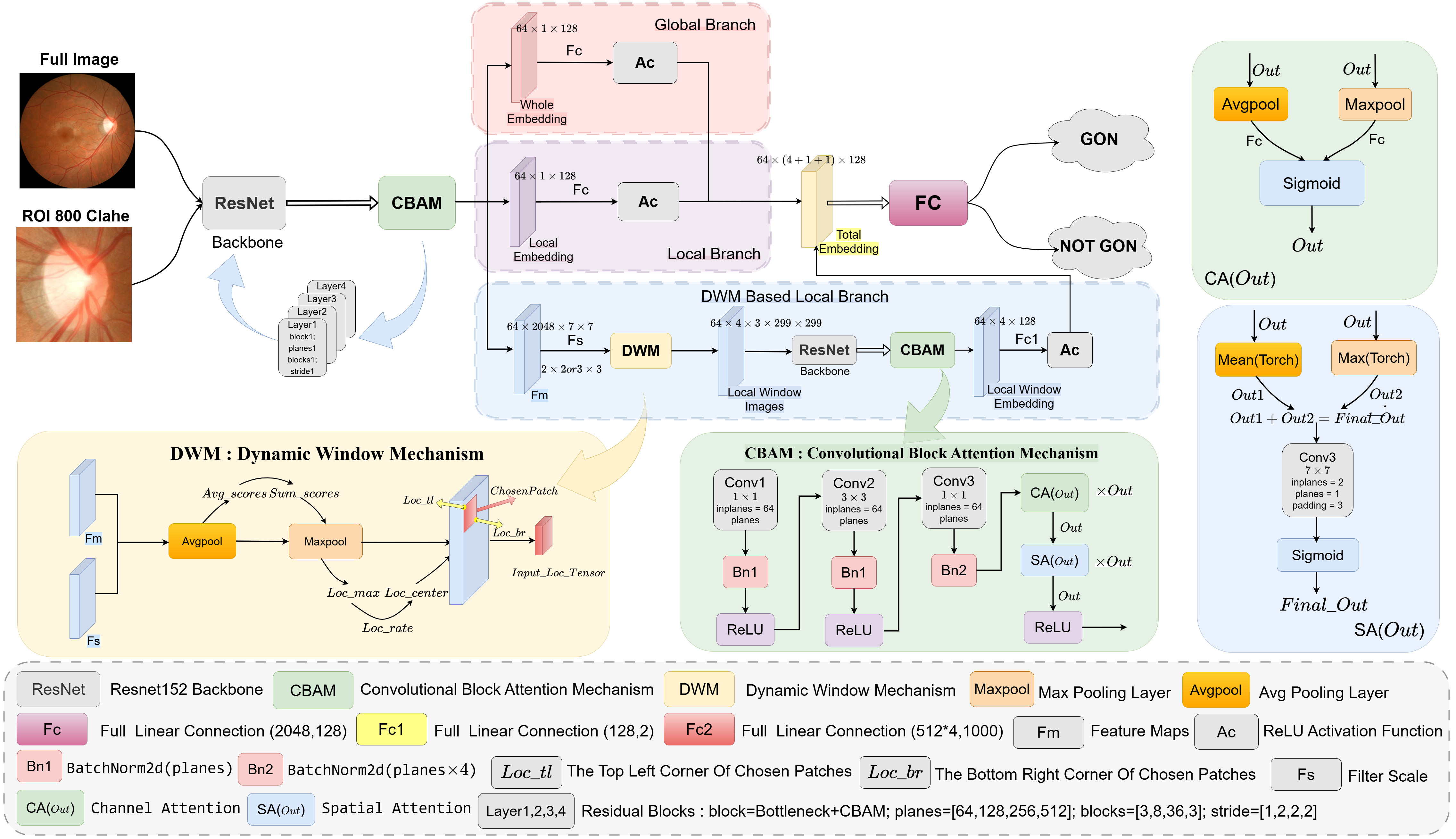}
    \caption{The architecture of our model. Global branch extracts global features and Local branch extracts subtle feature from local patches selected by \textbf{DWM}. The feature fusion method fuses the global feature embeddings and local feature embeddings for the classification.}
    \label{Fig.2}
\end{figure}

\cref{Fig.2} provides an overview of our cross-attention three-branch model, which processes retinal images as input. The Global Branch extracts spatial features from entire images, while the Local Branch consists of two sub-branches designed to extract local features from informative regions. To segment the ROI, we use a pretrained model from Fu et al.\cite{fu2018joint}. This pretrained model incorporates spatial constraints, equivalent augmentation, and cup proportion balancing, demonstrating a high performance in delineating the boundaries of OC and OD across a large-scale datasets.

Additionally, We employ a DWM to automatically select the optimal receptive regions, including the vasculature surrounding OC and OD. To further enhance feature extraction, we utilize ResNet as the backbone and optimize it with CBAM, naming our network as ResNet152-CBAM. The ROI and selected receptive regions are then passed to our two sub-branches which focus on local features and generate two local-feature embeddings. By integrating the global embeddings with the local embeddings from both sub-branches, the fused embeddings are delivered to the classier for the identification of 'referable' or 'non-referable' cases.

\subsection{ResNet152-CBAM}
Our network, Resnet152-CBAM, incorporates an attention mechanism to enhance feature extract by focusing on the most relevant information in both channel and spatial dimensions. To further improve the feature extraction, we integrate another CBAM as a post-process module to refine the extracted features.
Specifically, we introduce two additional attention mechanisms into the fundamental block of the ResNet backbone. 
The Channel Attention Mechanism highlights the most informative feature channels. It consists of a full connected layer that includes two convolutional layers and an activation layer. 
The first convolutional layer reduces the number of output channels while the second convolutional layer restores the number of input channels, thereby reducing computational complexity. The mechanism begins by applying adaptive maximum pooling and average pooling to the input feature map. 
The results of these pooling operations are then processed through the full connected layer, where they are integrated and normalized using the sigmoid function to generate channel-attention weights.

Spatial Attention Mechanism directs the network's attention to spatially important regions within the feature maps. It computes the maximum and average values along the channel dimension of the feature maps. These values are concatenated along an extra dimension and processed through a convolutional layer. The output is normalized using the sigmoid function to generate spatial attention weights. 
The attention weights derived from both mechanisms are then multiplied together, assigning varying importance to different regions the image in both the channel and spatial dimensions.

\subsection{Global Branch And Local Branch}
The global branch processes full images as input, focusing on extracting spatial features. This branch employs a convolutional feature extractor followed by four residual blocks. Each block is configured differently to process the feature maps through convolutions and downsampling operations while maintaining the consistent feature map dimensions throughout the branch. 
The output feature maps are denoted as $F \in \mathbb{R}^{C\times H\times W}$, where $C$ represents the number of channels and $H \times W$ denotes the spatial dimensions of the feature map. Then, the feature maps pass through an average pooling, followed by a fully connected layer, to generate the final global feature embeddings.

The second sub-branch, the local branch,  fully utilizes local features extracted from the ROI, which processes segmented ROI as input. The feature maps, after passing through the final convolutional layer, are represented as $F_1 \in \mathbb{R}^{C\times H\times W}$. Similar to the global branch, $F_1$ is further processed to generate local feature embeddings, encapsulating both spatial and structural information. Ultimately, the global and local embeddings are integrated with embeddings from the third branch to facilitate the downstream classification task.

\subsection{DWM Based Local Branch}
Our second sub-branch is responsible for extracting additional potential information from automatically selected regions. 
We design an automatic coordinating strategy that utilizes the feature maps $F$ of entire images to calculate a fixed number of locations containing the optimal receptive patches. 
The process begins by inputting $F$ into an average pooling operation, resulting in a score filter of size $H_f \times W_f$. 
Let $p$ denote the proposal size of each entire image; we then compute the size of score patches $H_s \times W_s$ and obtain the total score $S_{sum}$ by summing all the average scores $S_a$. 
To localize the center of each patch, we identify the maximum score $V_{max}$ within each patch and compute its location using $H_{max} = L_{flat}/W_s$ and $W_{max} = L_{flat} \bmod{W_s}$, where $L_{flat}$ denotes the index of $V_{max}$ in the flatten total score maps. 
The identified patch locations are then mapped back to the original entire image coordinates. 
The height $H_{loc}$ of each patch in the original image is computed by $H_{loc} = (2\times H_{max} + H - H_s + 1)/(2 \times H)$ while the width $W_{loc}$ is computed as $W_{loc} = (2\times W_{max} + W - W_s + 1)/(2 \times W)$. 

Next, to determine the optimal receptive regions, we prior compute loc\_max, the locations of the maximum values of the total scores within the flatten images. Then, let fm\_h denote the height of image maps, fm\_w denote the width, H and W respectively denote the height and width of images, the height rate \textbf{loc\_rate\_h} and the width rate \textbf{loc\_rate\_w} are calculated as follows. The central locations \textbf{loc\_center} are computed based on loc\_rate\_h and loc\_rate\_w. 
After compute the rate of height and width, let $H_p$ and $W_p$ denote the size of corresponding patches. 
With the size of patches, the top left corner \textbf{loc\_tl} and the bottom right corner \textbf{loc\_br} of chosen patches will be calculated by following two functions. 
Based on the determined center and four corners of the optimal receptive patches within the entire images, the target patches are segmented out and the most relevant patches are selected by ranking the scores. 
These selected patches are then reshaped to an adaptive size and fed into our network. The resulting output feature maps are denoted as $\sum_{i=1}^{p}{F_{2i}} \in \mathbb{R}^{C \times H\times W}$. These feature maps are processed through average pooling followed by a fully connected layer to obtain the second set of local feature embeddings. 
Finally, the embeddings are integrated with the global and local embeddings, and the final feature embeddings are delivered to the downstream classifier for the final prediction. 

\begin{equation}
loc\_rate\_h = \frac{2 \times loc\_max[:, 0] + fm\_h - H + 1}{2 \times fm\_h}
\end{equation}
\begin{equation}
loc\_rate\_w = \frac{2 \times loc\_max[:, 1] + fm\_w - W + 1}{2 \times fm\_w}
\end{equation}
\begin{equation}
loc\_tl =\left( loc\_center[:, 0] - \frac{H_p}{2},loc\_center[:, 1] - \frac{W_p}{2} \right)
\end{equation}
\begin{equation}
loc\_br = \left( loc\_center[:, 0] + \frac{H_p}{2} + (H_p \% 2),loc\_center[:, 1] + \frac{W_p}{2} + (W_p \% 2) \right)
\end{equation}


\section{EXPERIMENTS}
\subsection{Experiment Environment}


\subsubsection{Datasets}
\begin{figure}[t]
    \centering
    \includegraphics[width=0.83\linewidth]{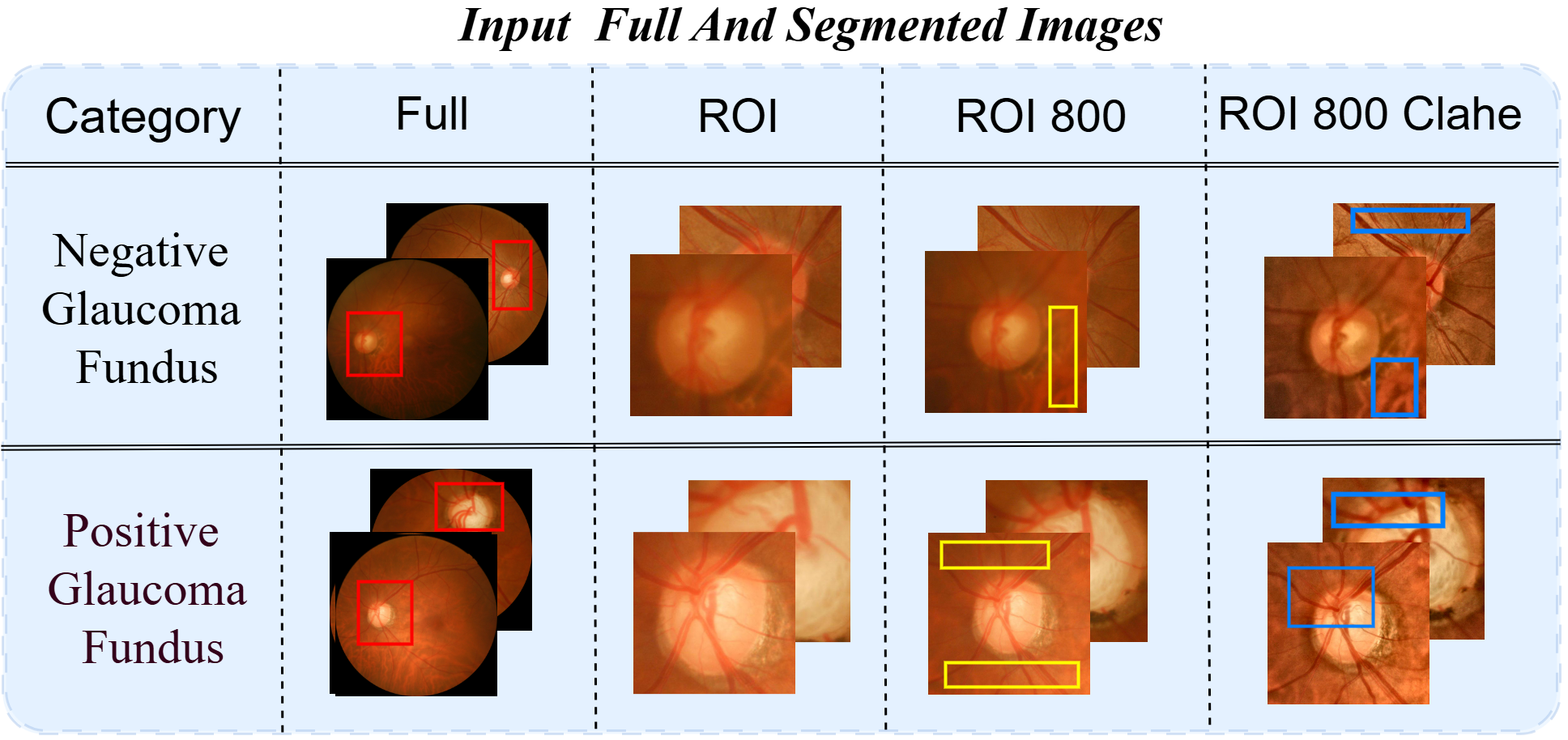}
    \caption{The samples both contain full images,segmented ROI, ROI 800 (in a higher view) and \textbf{ROI 800 Clahe (Contrast Limited Adaptive Histogram Equalization)} which is our final input. \textbf{Red boxes} mark the target region ROI,\textbf{Yellow boxes} are used to mark the effect of higher view and \textbf{Blue boxes} mark the improving image contrast.}
    \label{Fig.1}
\end{figure}

The dataset used in our experiment is the Rotterdam EyePACS AIROGS dataset \cite{r31}, which contains a large collection of color fundus images from diverse subjects across multiple sites, representing a heterogeneous ethnic population. Our training dataset consists of 36,803 images, including 9,284 referable and 27,519 non-referable samples. The testing dataset comprises 1,999 images, with 488 referable and 1,511 non-referable samples. Additionally, we segment ROI from the full images to create a separate dataset, as illustrated in \cref{Fig.1}. During the training and testing process, ROI 800 Clahe is selected as the final input, as it demonstrates superior performance in enhancing ROI features by improving image contrast. This preprocessing step helps mitigate variations caused by differences in equipment or environmental factors, thereby playing a crucial role in improving the overall performance of our model.

\subsubsection{Model Structures}

We design three comparative neural network for this binary classification task:
\begin{enumerate}
    \item Patch5Model: A three - branch model that uses ResNet152 backbone.
    \item Branch2CBAM: A cross - attention two - branch model that utilizes ResNet152 equipped with CBAM.
    \item Branch3CBAM: A cross - attention three - branch model based on ResNet152 backbone with CBAM.
\end{enumerate}

The full and ROI images are loaded as model input in [64, 6, 3, 224, 224]. To simulate real-world conditions and enhance model robustness, we apply random horizontal flip, vertical flip, color jitter, and Gaussian blur during training. These augmentations expose the model to diverse fundus images. The enhanced data is then normalized and resized to [64, 6, 3, 299, 299] to meet Branch3CBAM input requirements.
In the network, full and ROI data are processed into feature maps of size [64, 2048, 7, 7], with output embeddings of [64, 2048]. The local window regions in the third branch are determined based on these feature maps. Full images and filter scales $S$ = [3,3] and [2,2] are used to compute optimal receptive regions, with patch sizes in [224, 224] and [112, 112]. Average scores are computed in [7, 2048, 5, 5] and [7, 2048, 6, 6] using avg\_pool2d, with total scores sized [7, 1, 5, 5] and [7, 1, 6, 6].


W and H are respectively in choice of [5,6]. After flatting the total scores into the size of [7, $W\times H$], the maximum value and index of the total scores are processed into maximum locations in size of [7, 2]. Then the total scores are delivered into max\_pool2d, the maximum pool of torch functional module. To localize the optimal receptive patches, we compute the height and width rates of receptive patches to the entire images. After computing the center of the receptive patches, we calculated the indices of the top-left corner and bottom-right corner within the flatten entire images. By combining the former two corner vectors, the optimal receptive regions can be obtained and the final shape of the input location tensors are [7, 6, 4].

The local window images, obtained based on the calculated locations within full images, are embedded into local window embeddings in size of [$64\times 4$, 3, 299, 299]. Ultimately, all the embeddings are fused into the final embeddings in size of [$64 \times (4+1+1)$, 128] which are fed into a full connected layer with the output of the final classification task.

\subsubsection{Parameter Settings}
The optimizer parameters are configured as follows: the initial learning rate is 0.001 for the first 10,000 iterations. The Adam optimizer uses a momentum term of 0.9 and a batch size of 64. For logging and checkpointing, the loss is recorded in TensorBoard every 400 iterations. Results are saved every 2,000 iterations, and checkpoints are stored every 10 epochs, with training resuming from the latest checkpoint starting at epoch 1.

\subsection{Evaluation Metrics}
To compare the performance of our cross-attention three-branch model with Patch5Model and Branch2CBAM, we use the following evaluation metrics: 1) average precision (AP): measures the ability to distinguish referable from non-referable images. 2) area under the curve (AUC): evaluates classification performance across various sensitivity-specificity trade-offs. 3) accuracy ($Acc$): represents overall classification correctness between referable and non-referable samples. 4) sensitivity ($Sen$): assesses the model’s ability to identify referable samples. 5) specificity ($Spe$): measures the model’s ability to correctly classify non-referable samples.. 6) F1-score ($F1$): balances precision and recall, providing a trade-off between sensitivity and specificity.

\subsection{Result Analysis}

\begin{figure}[ht]
    \centering
    \includegraphics[width=1\linewidth]{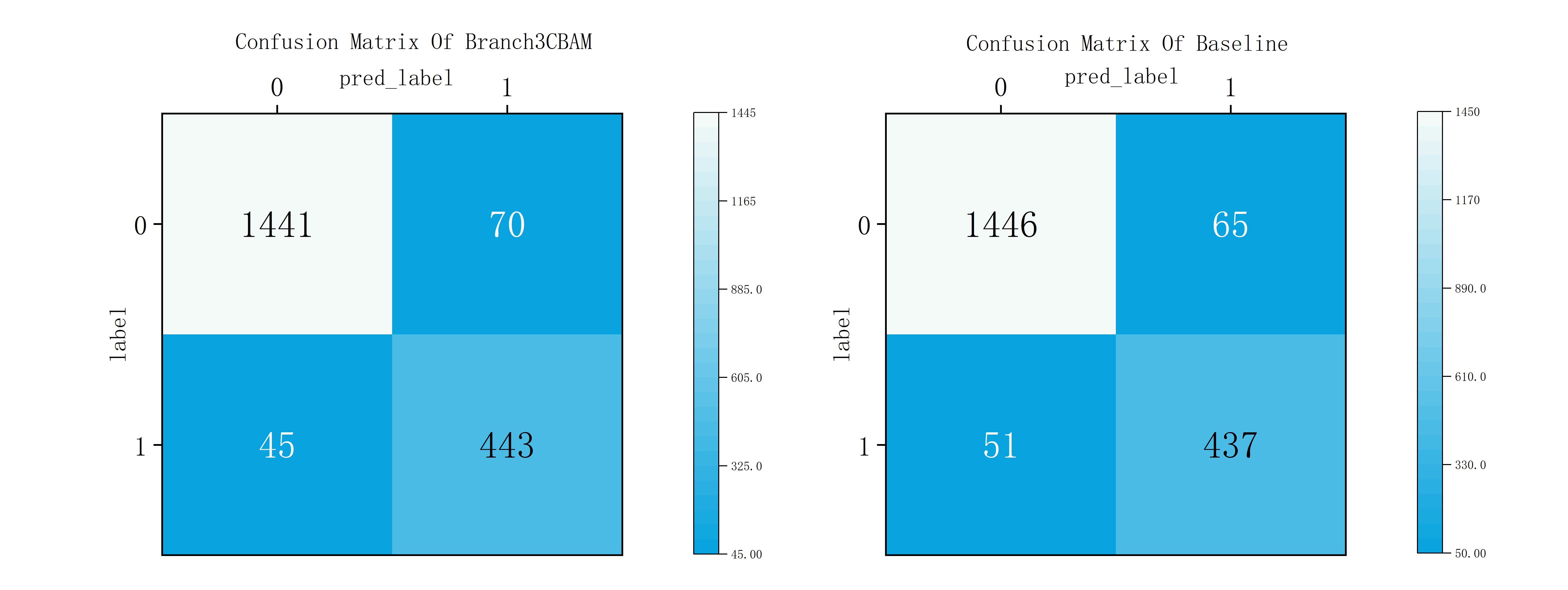}
    \caption{The predicted results of our proposed model.}
    \label{fig4}
\end{figure}

\cref{fig4} illustrates the exact performances of both Branch3CBAM and the baseline model in the classification tasks, presenting the detailed confusion matrix and highlighting the TP, TN, FP, FN. Branch3CBAM achieves higher accuracy in identifying referable samples, but makes slightly more mistakes while identifying non-referable samples. This trade-off in Branch3CBAM performance may be given rise to the employed cross-attention mechanism compared to the baseline. The increased complexity of model architecture enables the model to capture more potential features from referable samples while simultaneously introducing more noise when processing non-referable images.

\begin{table}[hb]
\centering
\caption{Evaluation Metrics For Comparative Methods}
\label{tab1}
\setlength{\tabcolsep}{8pt} 
\begin{tabular}{c c c c c c c}
\hline
\multirow{2}*{\centering \textbf{Models}}  & \multicolumn{6}{c }{\textbf{Evaluation Metrics}}  \\ \cline{2-7}  
                        & AP  & AUC    & Acc   & F1 & Sen & Spe    \\ \hline
Patch5Model      & 0.954 & 0.982 & 0.941 & 0.883 & 0.895
 & 0.957  \\ \hline
Branch2CBAM       & 0.947 & 0.981 & 0.928 & 0.866 & 0.957 & 0.919  \\ \hline
Branch3CBAM   &  0.955 & 0.982  & 0.942 & 0.885 & 0.908 & 0.954 \\ \hline

\end{tabular}
\end{table}

To ensure an optimal layout, the evaluation metrics for the comparative methods are detailed in \cref{tab1}, providing a quantitative assessment of their classification performances. 1) $AP$: Branch3CBAM achieves an $AP$ of 0.955, which is slightly higher than Patch5Model's 0.954 and much higher than Branch2CBAM's 0.947, indicating that Branch3CBAM is better at identifying positive samples. 2) $AUC$: Branch3CBAM achieves an $AUC$ of 0.982446, which is almost identical to Patch5Model's AUC 0.982417 while Branch3CBAM outperforms Branch2CBAM (0.981). This indicates that the third branch using DWM further enhances the model’s discriminative power. 3) $Acc$: Branch3CBAM has an $Acc$ of 0.942, higher than both Patch5Model (0.941) and Branch2CBAM (0.928), showing that Branch3CBAM retains high accuracy while benefiting from the model complexity and deeper feature extracting method. 4) $F1$: Branch3CBAM achieves an $F1$ score of 0.885, slightly higher than Patch5Model (0.883) and much better than Branch2CBAM (0.866), highlighting Branch3CBAM's superior performance in balancing $precision$ and $recall$. 5) $Sen$: Branch3CBAM’s $Sen$ (0.908) is better than Patch5Model (0.895). However, the sensitivity of Branch3CBAM is lower than Branch2CBAM (0.957). This is likely due to the increased complexity of model architecture which might introduce slight over-fitting to dominant pattern, leading to the reducing sensitivity. 6) $Spe$: Branch3CBAM achieves a specificity of 0.954, surpassing Branch2CBAM (0.919), but slightly lower than Patch5Model (0.957). This marginally lower specificity compared to Patch5Model may result from CBAM which enhances subtle together with less discriminate features, leading to an increased false positive rate. Nevertheless, compared to Branch2CBAM, the introduction of our third sub-branch in application of DWM allows better negative sample recognition.


\begin{figure}
    \centering
    \includegraphics[width=1 \linewidth]{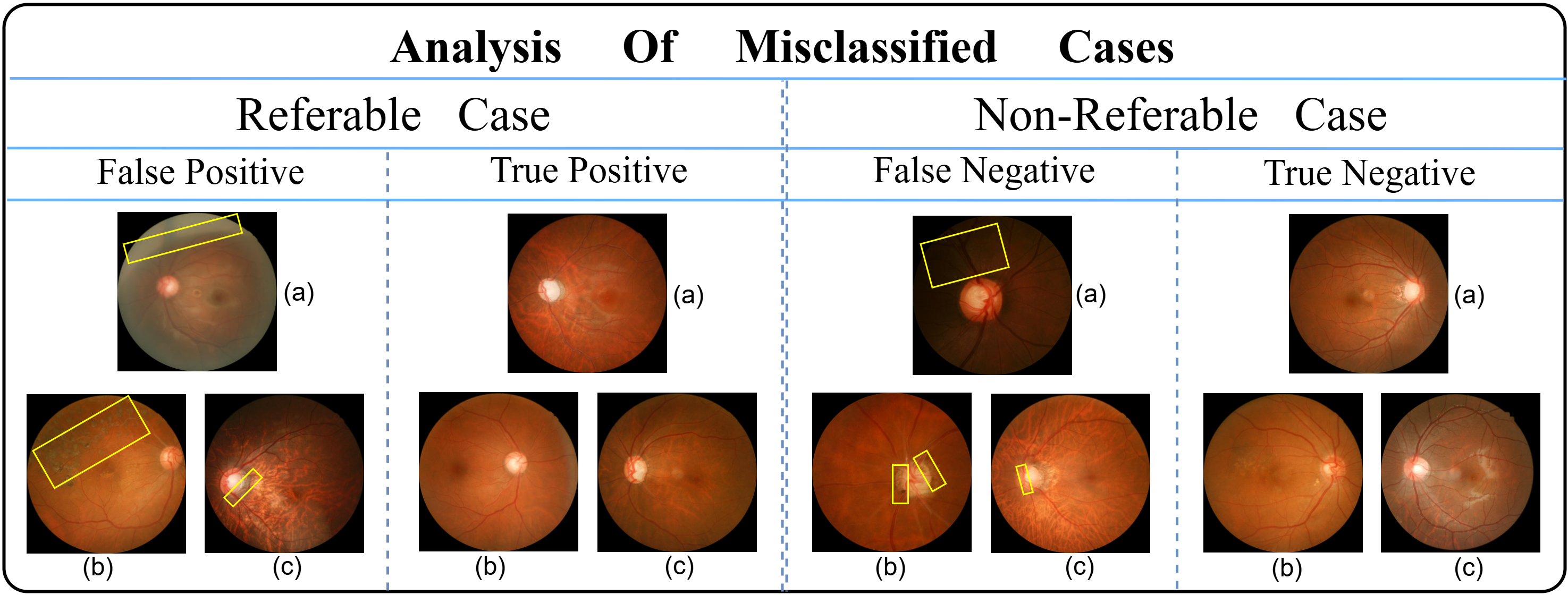}
    \caption{Figure 5 illustrates the misclassifications in both 'referable' and 'non-referable' categories. \textbf{Yellow boxes} highlight the regions where various factors may misdirect the model to make errors potentially.}
    \label{fig5}
\end{figure}

To better understand the sources of misclassification in distinguishing between 'referable' and 'non-referable' cases, we present misclassified samples alongside correctly classified ones for comparison in \cref{fig5}.  
For the referable cases, misclassification in Figure (a) may be attributed to poor exposure and low resolution, while Figure (b) might be affected by noise from other ocular diseases. Figure (c) demonstrates an uncertain optic cup-disc boundary, which could have led to misprediction. 
In the non-referable cases, the misclassification in Figure (a) is likely due to the insufficient highlight in the picture. Figure (b) also exhibits an ambiguous cup-disc boundary, similar to the issue in referable cases. Figure (c) may be misclassified due to its low resolution in imaging the small blood vessels while the correctly classifies samples in the rightmost column all have clear and well-defined vessel structures.

\begin{table}[h]
\centering
\caption{Performance comparison with the latest detection methods}
\label{tab2}
\begin{tabular}{c c c c c c}
\hline
\textbf{Author} & \textbf{Method} &
Acc    &      Sen     &      Spe      &      AUC   \\ \hline
Gomez et al. \cite{r16}     & VGG19 TL                    & 88.05 & 87.01 & 89.01 & 94 \\ \hline
Diaz et al. \cite{r17}       & Pre-trained CNN model: Xception       & 89.77 & 93.46 & 85.80 & 96.05  \\ \hline
Liu et al. \cite{r32}  & Deep CNN architecture                    & 92.7 & 87.9 & 96.5 & 97  \\ \hline
Gheisari et al. \cite{r19} &    VGG16 with Long Short-Term Memory    & - & 95 & 96 & 99  \\ \hline
Proposed Method & Resnet152-CBAM-3B with DWM  & 94.25 & 90.78 & 95.37 & 98.24  \\ \hline
\end{tabular}
\end{table}

Table \ref{tab2} presents a comparison of the latest deep - learning - based methods for glaucoma detection on public and private datasets. To compare the performance of all the latest methods, we select the evaluation metrics, including accuracy, sensitivity, specificity and AUC. 
Notably, out method achieves the highest accuracy among all the methods, indicating the superior overall performance of our proposed method in glaucoma detection. 
Compared to VGG19 TL proposed by Gomez et al. \cite{r16}, our proposed method surpasses their method in all four metrics. 
For Diaz et al. \cite{r17} who utilzies the CNN method : Xception, our method achieves slight low sensitivity (90.78 vs 93.46), generating more wrongs while missing 'referable' samples. However, our method compensates with higher accuracy, AUC and especially higher specificity. 
A recent study by Liu et al.\cite{r32}, using a deep CNN architecture, showed a 96.5 in specificity in glaucomatous disc identification which shows a slightly better ability of identifying 'non-referable' cases than our method. Nevertheless, in the terms of accuracy, sensitivity and AUC, our methods outperforms their method with conspicuous improvement. 
Gheisari et al. \cite{r19} employs VGG16 with a designed memory mechanism, demonstrating great sensitivity, specificity, and AUC. However, since its accuracy is not explicitly evaluated, its overall reliability remains uncertain. 
With a well-balanced performance across all metrics while all the scores surpass 90, our proposed method, ResNet152-CBAM-3B with DWM, achieves remarkable harmonization and stability, ensuring its reliability in glaucoma detection. 

\section{CONCLUSION}

In this work, we propose a cross-attention three-branch model that integrates the CBAM and DWM. The third branch, introducing DWM, complements the global and local branches, which are limited by their fixed focus on specific regions, thereby addressing uncertainty in cup-disc boundary determination. CBAM enhances the network by incorporating channel and spatial attention mechanisms, enabling adaptation to complex retinal imaging scenarios. Experimental results demonstrate that Branch3CBAM outperforms other models with greater stability and higher accuracy, even when handling images of varying resolutions and inconsistent quality from multiple imaging devices. For future work, we aim to develop a more generalizable module to enhance efficiency and accuracy in glaucoma detection, particularly for datasets representing diverse racial and geographical populations.

%
%
%
%

\bibliography{sample}
\bibliographystyle{splncs04}

\end{document}